\documentclass{article}

\usepackage[final,nonatbib]{icbinb2021}

\usepackage[utf8]{inputenc} 
\usepackage[T1]{fontenc}    
\usepackage{hyperref}       
\usepackage{url}            
\usepackage{booktabs}       
\usepackage{amsfonts}       
\usepackage{nicefrac}       
\usepackage{microtype}      
\usepackage{xcolor}         
\usepackage[pdftex]{graphicx}
\usepackage{color}
\usepackage{subcaption}

\newcommand{\hush}[1]{}

\title{Text Ranking and Classification \\ using Data Compression}

\author{%
  Nitya Kasturi\\
  Meta Platforms Inc \\
  nityakasturi@fb.com \\
   \And
   Igor L. Markov \\
   Meta Platforms Inc \\
   imarkov@fb.com \\
}

\begin{document}

\maketitle

\begin{abstract}
  A well-known but rarely used approach to text categorization uses conditional entropy estimates computed using data compression tools. Text affinity scores derived from compressed sizes can be used for classification and ranking tasks, but their success depends on the compression tools used. We use the {\tt Zstandard} compressor and strengthen these ideas in several ways, calling the resulting
  language-agnostic technique {\tt Zest}. In applications, this approach simplifies configuration, avoiding careful feature extraction and large ML models.
  Our ablation studies confirm the value of individual enhancements we introduce.
  We show that {\tt Zest} complements and can compete with language-specific multidimensional content embeddings in production, but cannot outperform other counting methods on public datasets. 
\end{abstract}

\section{Motivation}
\label{sec:motivation}

The idea of comparing texts using off-the-shelf lossless data compression tools goes back to \cite{LangTrees}, which in 2002 linked {\em entropy estimation} and using gzip on text with text similarity metrics. Given two strings $X$ and $Y$, one compresses each of them individually and also the string $X + Y$. Similar strings compress better after being concatenated. An affinity score for $X$ and $Y$ is computed from the three resulting bytesizes. This computation is simple, requires little infrastructure, works for any language, and naturally handles similar words, word forms, typos, etc. It can be easily applied to multi-class classification (e.g., binning news articles by category) and ranking relative to known examples.

Following up on \cite{LangTrees}, the 2004 work in \cite{DataMining} showcased the versatility of compression-based estimators
by demonstrating their practical advantages for (i) clustering time-series and text, (ii) anomaly detection, (iii) classification for various non-text datasets. These techniques are attractive in practice because they can handle many data formats without the need to understand their structure, are easy to configure, and are reasonably efficient in terms of computation. The authors of \cite{DataMining} also emphasize the ``parameter-free'' nature of compression-based estimators, in comparison with various ML techniques that may produce different results depending on parameter settings. In practice, some compression tools are better than others on certain kinds of inputs (making a choice of a compression tool is a type of fitting to data), and hyperparameter settings affect compression performance: compression block size and word size (8/12/16 bits), level of effort, parallel execution, efficient use of CPU instruction sets, etc. Moreover, compression headers and dictionaries embedded in each compressed file spoil entropy estimates, especially for small files. In fact, a controversy ensued after the publication of \cite{LangTrees} because zipping could not outperform Naive Bayes~\cite{LangTeesResponse}. 

Despite the past controversies and apparent lack of broad adoption,
the idea to use data compression for text classification has become
so mainstream today that it appears in a key AI textbook \cite[Chapter 23]{AItextbook}. There, Russel and Norvig point out that "In effect, compression algorithms are creating a language model. The LZW algorithm in particular directly models a maximum-entropy probability distribution." Presumably, compression ratio then estimates the probability that given input was generated by the language model.
Motivated by this discussion, Halford offers~\cite{DataCompression} are a recent (2021) re-evaluation of the idea from \cite{LangTrees}, using a recent version of {\tt gzip} compressor on the public News data set available with {\tt sklearn} \cite{NewsDataset}. Indeed, Multinomial Naive Bayes from {\tt sklearn}  clearly outperforms {\tt gzip}-based classification by F1 score and "It’s also dramatically faster". Halford evaluates several other modern compressions --- {\tt zlib}, {\tt bz2}, and {\tt lzma}, which show wildly different results in terms of F1 and runtime. All of these tools are much slower than Multinomial Naive Bayes, and only {\tt lzma} shows performance on par with it but at the cost of being several times slower than other data compressors. In contrast to these recent pessimistic results, we show how to significantly improve the ML performance and runtime of compression-based text classification.

\section{Insights and Overview}
We revisit compression-based text affinity scores \cite{LangTrees}  because modern compressors are faster than tools from 2002 and produce better entropy estimates. The lossless open-source {\tt zstandard} compressor \cite{zstd,YCollet2016,RFC}, developed by Yann Collet at Facebook around 2016, produces results not far from those of (slow) arithmetic coding that are considered close to the Shannon bound. The {\tt zstandard} compressor offers a {\em dictionary interface}, which allows us to improve the approach of \cite{LangTrees} and make it more practical, especially for small and medium-sized strings. {\tt Zstandard} can build a compression dictionary and use it to compress many small files, to avoid the overhead of separate dictionaries. We use the dictionary interface to compress texts to be classified or ranked. This sharpens entropy estimates and improves speed versus compressing concatenated files (where the same file would be concatenated with multiple other files).
The resulting text affinity scores can be used as inputs to multimodal classifiers and rankers.
The simplicity of this language-agnostic method is attractive when building ML platforms, especially for product engineers without ML background.\hush{, while highly optimized data compression codes are compute-efficient. In our experiments, using simple frequency-based (n-gram) character-level language models produced slightly more accurate results than compression-based methods, but the gap in efficiency was significant.} 


We leverage text affinity estimation in text ranking and classification. For example, given positive and negative examples for 2-class classification, we first build compression dictionaries for each class. As an option, the texts can be normalized by removing punctuation within sentences (but not spaces) and lowercasing the remaining letters (for languages without upper/lower cases, this is a no-op). With
{\tt Zstandard}, there is no need to concatenate files as in \cite{LangTrees}, thanks to the dictionary interface.

The original approach has several major weaknesses that we are able to address. To illustrate them, consider classifying news articles in topics --- Politics, Celebrities, and Sports.  Some important words and phrases appear in multiple topics, but with different frequencies, for example, “Arnold Schwarzenegger”. For a sufficiently large set of examples, such words compress equally well for each class, and do not contribute useful information. This is particularly detrimental when classifying or ranking short texts. Downsampling the examples would help with common words and phrases, but undermine the handling of rare words and phrases (which would not be compressed for any class). To address this challenge, we use {\em a set of dictionaries of telescoping sizes} --- this way, common words are differentiated by smaller dictionaries and rare words are differentiated by larger dictionaries. Another challenge is the heavier impact of longer words on compression ratios. We address it by {\em word padding} to fixed length, e.g., "hello" padded to 10 characters becomes "hellohello".
We configure {\tt zstandard} to minimize headers in compressed files and, furthermore, subtract the compressed size of an empty string from compressed sizes of evaluated pieces of text. For each evaluated text, for each classification class, we average the byte compression ratio over multiple dictionaries. Subtracting this number from 1.0 produces an affinity score, for which “greater is better”. In particular, a sentence that was seen in some class examples may return a value close to 1.0, whereas a sentence in a different script (e.g., Greek vs Cyrillic) would not compress well, resulting in a value close to 0.0. For multiclass classification, we subtract the min class score from all scores. This handles words present in many classes. For ranking applications, affinity scores can be sorted to produce an ordering.

\section{Implementation and Empirical Evaluation}
\label{sec:evaluation}

Our {\tt PyTorch} implementation is based on an untrained {\tt Torchscript} module for {\tt Zest}. The Python {\tt zstandard} library was not supported with {\tt TorchScript}, so we implemented the {\tt Zstandard} interface in C++ as a {\tt TorchScript} module using the original {\tt C} library. For comparision with other linear models and public text datasets, we used the Python {\tt zstd} library directly.
The {\tt Zest} transformer takes in lists of text containing the examples per class to train separate dictionaries, as well as a list of text features for evaluation. It produces affinity scores per class, which can be used separately or combined (e.g., by subtraction) into one score.

\subsection{Deployment in a production ML Platform}

We onboarded {\tt Zest} to a company-internal ML platform that hosts hundreds of ML models for prediction, classification, etc. Using production data, we evaluate the use of {\tt Zest} text affinity scores as additional features to these models, where we can judge features by their importance values in the context of various other features. The specific application discussed in this paper uses text features to rank search results in an internal tool (posts) produced for a given search query. Some model features provide context: {\tt user\_id}, {\tt group\_id}, and post {\tt owner\_id}. The most useful features in the deployed ML model include the number of characters, query proximity to the post text, and components of a multi-dimensional content embedding. In this study, we check if compression-based affinity scores are comparable in their utility to language-specific content embeddings, which require additional training and cannot handle text in many languages, or mixed-language text. In contrast, {\tt Zest} can be useful where content embeddings are not easily available.

 To prepare input for text affinity computation, we feature-engineer the positive and negative post examples to evaluate the post text. For each user on the internal tool, we fetch the 12 most recent posts from the 5 groups with which the user interacted most recently. We split these examples between positive and negative by whether the user viewed them more than once. As most posts end up being negative examples for a user, we added posts that are trending on the internal tool as positive examples. 
On average, a user has 39 negative examples with 851 characters each and 20 positive examples with 1659 characters each. This way, the number of characters passed into the positive and negative compression dictionaries balances out.

\subsection{Empirical evaluation within a larger model}

We ran a baseline gradient-boosted decision tree (GBDT) workflow with no additional features, a GBDT workflow with {\tt Zest} scores as features, and a GBDT workflow with 3-gram affinity scores (percentage of matched 3-grams) as features. 
The {\tt Zest} transformer took roughly 50\% longer to compute affinity scores in our workflows compared to the n-gram transformer (both implemented by us). However, the {\tt Zest} classifier code ran much faster with the Python implementation, taking 200-300 seconds to build dictionaries on 25k examples, whereas the transformer would take 40-50 minutes for a few thousands of examples. This could be to due to the memory limits placed on TorchScript modules when running in production. Feature importance of the top Zest feature was \#16 (compared to \#201 for the n-gram models), ahead of hundreds of content-embedding dimensions and behind of only 7 of them.

We then tested removing embedding features to see how well {\tt Zest} can fill in the missing information. Based on random removal of embedding dimensions, the new features allow the model to drop 150 dimensions of both the post and query embeddings, while improving normalized entropy (NE) by 7\% compared to the baseline model. With half of the embedding features removed, the top {\tt Zest} feature was at \#8 with only 4 embedding features ahead. An important distinction between {\tt Zest} and the embedding features is that the Texas SIF embeddings \cite{SIF-2017, SIF-2018} used in this application can only support English and Spanish, whereas the {\tt Zest} transformer is language-agnostic and can handle mixed-language text. Incidentally, the Texas SIF embeddings are shown be a "Tough-to-Beat Baseline for Sentence Embeddings" \cite{SIF-2017}.
\hush{Although in training it is easy to remove embedding dimensions from a feature group, feature extraction costs and performance overhead remain the same if we choose to use only certain features from a group in WWW. To unlock resource savings, smaller feature groups can be generated or embeddings can be trained with fewer dimensions to save resources.}
We also checked feature quality of the {\tt Zest} transformer by removing features representing an ID, given that the ID features had high feature importance and could be affecting model quality. After removing the ID features, feature importance jumped from \#16 to \#9, again being outperformed by only 4 embedding dimensions.

Table~\ref{table:production} compares different workflow runs with different configurations along with the offline results for comparison.
There is a distinction in the score distribution between the True and False examples, with most True examples around 0.03 - 0.11, and most False examples around 0-0.07, indicating that the feature values can be strong enough to help with classification on its own.

\begin{figure}
  \centering
  \begin{subfigure}[b]{0.45\textwidth}
    \centering
    \includegraphics[width=\textwidth]{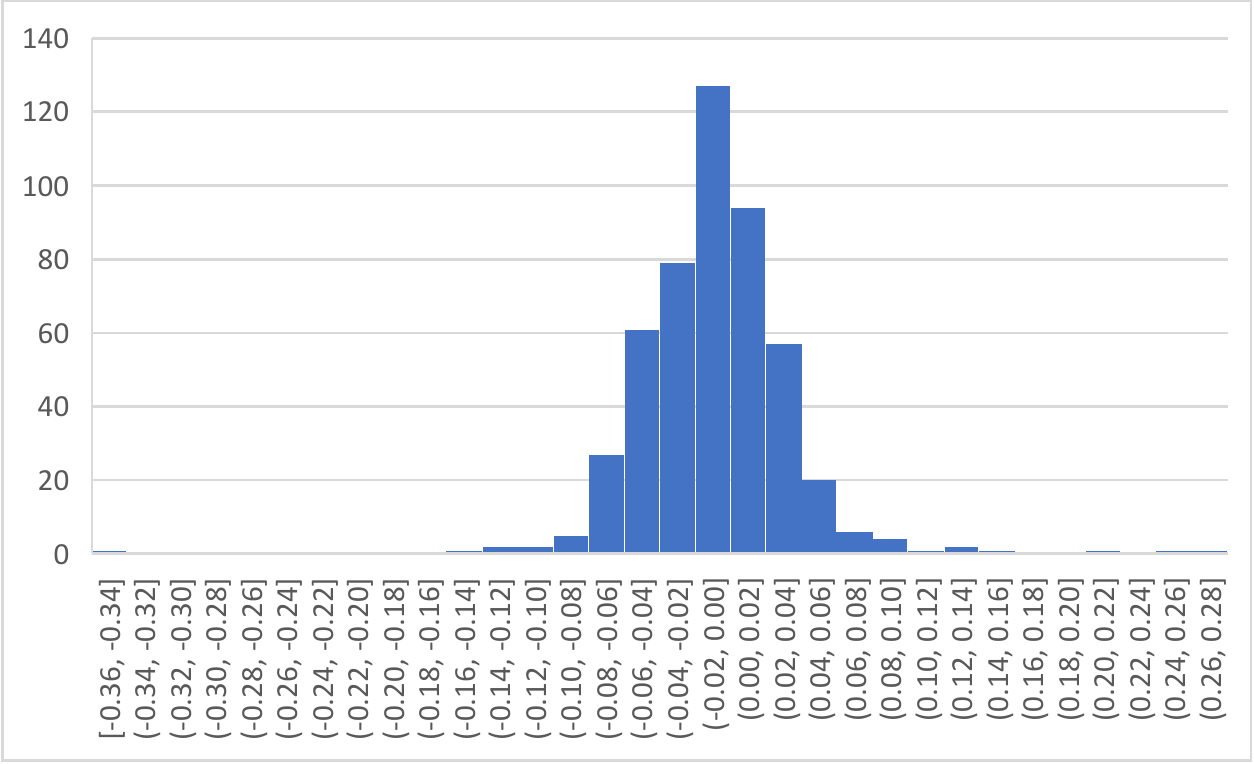}
  \end{subfigure}
  \begin{subfigure}[b]{0.45\textwidth}
    \centering
    \includegraphics[width=\textwidth]{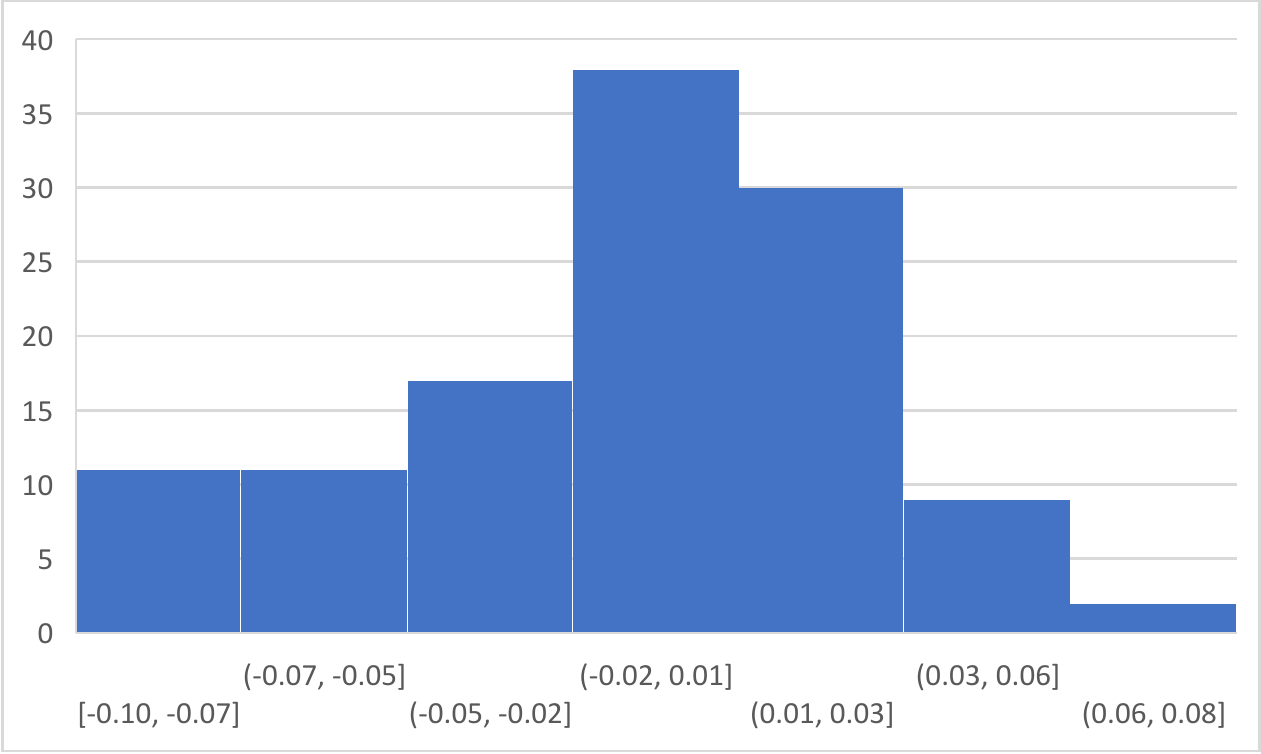}
  \end{subfigure}
  \caption{{\tt Zest} score distribution for True and False examples respectively.}
\end{figure}

\hush{When extracted in WWW, the Texas SIF embeddings float features need Predictor to call a trained model to generate embeddings for a text sample (done twice for the body and the query), taking an average of 0.03 seconds per example, not including training time. In comparison, the {\tt Zest} transformer in C++ takes 6 $ms$ per example (the Python implementation runs much faster, which hints at fixable inefficiencies in the C++ interface, since {\tt Zstandard} is written in C).}

\subsection{Standalone comparisons versus linear models}

As demonstrated on production data, Zest features can be useful when passed into a model along with other features.
To compare {\tt Zest} to other text transformers while
avoiding non-text features, to perform ablation studies, and experiment with ensembling, we worked with various public text classification and sentiment analysis datasets. We ran Logistic Regression (LR) trained on Bag of Words (BoW) as a feature, LR trained on Facebook's {\tt InferSent} sentence embeddings \cite{InferSent}, and the multi-class version of {\tt Zest} on a dataset with various categories of news headlines \cite{NewsDataset} and various sentiment datasets (Stanford Movie sentiment \cite{StanfordMovie} and IMDB movie review sentiment \cite{IMDBMovie}).
We ran {\tt Zest} with 1, 2, and 4 telescoping dictionaries each to check performance on the news headlines dataset. The {\tt Zest} model with 4 telescoping dictionaries performed significantly better than the rest. Word padding generally improved accuracy by 0.5-1\% based on the dataset. Compared to sophisticated language models --- BERT and such, --- {\tt Zest} has a much smaller resource footprint and is easier to work with, yet customizable. 

The ensemble (averaged) prediction of {\tt Zest} and LogisticRegression with a BoW performs the best on the news headline categories dataset. However, BoW as a feature worked well with Logistic Regression --- it ran faster and outperformed standalone {\tt Zest} by 1-5\% in accuracy, depending on the dataset. The sentiment analysis datasets both had an accuracy of $\approx$87\% with LR and BoW versus $\approx$78-80\% for Zest, no matter how it was ensembled. 

\begin{table}[h]

\begin{minipage}{0.51\textwidth}
        \centering
        \caption{Comparisons in a production setting}
        \label{table:production}
        \begin{tabular}{l @{\hspace{0.75\tabcolsep}} l @{\hspace{0.75\tabcolsep}} l @{\hspace{0.75\tabcolsep}} l @{\hspace{0.75\tabcolsep}} l}
            \toprule
            model  & \small{post emb.} & \small{query emb.} & \small{feat. imp.} & NE \\
            \midrule
            \small{Baseline} & All  & All  & ---              & \small{0.196}  \\
            {\tt Zest}       & All  & All  & 0.29\%           & \small{0.212}  \\
            3-gram           & All  & All  &  0.06\%          & \small{0.216}  \\
            {\tt Zest}       & None & None &  \textbf{3.57\%} & \small{0.448}  \\
            {\tt Zest}  & \small{150 dims} & \small{150 dims} & 1.54\% & \small{0.189}  \\
            \small{Baseline}       & All & All &  --- & \small{0.173}  \\
            \small{Baseline}        & None & All & --- & \small{0.161}  \\
            {\tt Zest}        & None & All &  \textbf{1.14\%} & \small{\textbf{0.151}}  \\
            3-gram        & None & All &  0.17\%  &  \small{0.151} \\
            \bottomrule
        \end{tabular}
\end{minipage}
\begin{minipage}{0.49\textwidth}
    \centering
    \caption{Comparisons for news headlines \small{\cite{NewsDataset}}}
  \begin{tabular}{l@{\hspace{0.75\tabcolsep}} l @{\hspace{0.75\tabcolsep}} l @{\hspace{0.75\tabcolsep}} l}
    \toprule
    model     & \small{size (MB)}  & \small{train sec.} &  acc. \\
    \midrule
    NB/BoW & 3.34  & 0.170 &  \small{0.926}  \\
    LR/BoW & 1.67 & 26.53 & \small{\textbf{0.947}}   \\
    LR/InferSent   & 777 & \textbf{822.3} & \small{0.874} \\
    {\tt Zest} 4D    & 3.04   & 74.43 & \small{0.924} \\
    \small{{\tt Zest} 4D/LR-BoW}   & 4.71  & 100.9 &  \small{\textbf{0.951}} \\
    {\tt Zest} 2D & 1.02  & 60.49 &  \small{0.886}   \\
    \small{{\tt Zest} 2D/LR-BoW}    & 2.69  & 87.02 & \small{0.946} \\
    {\tt Zest} 1D   & 0.02       & 49.83 & \small{0.741} \\
    \small{{\tt Zest} 1D/LR-BoW}  & 1.69      & 76.36 & \small{0.938} \\
    \bottomrule
  \end{tabular}
    
\end{minipage}
\vspace{1em}
\caption*{The baseline model is a GBDT. Alternatively, we pass either {\tt Zest} or the 3-gram counter in as additional features to the baseline model. BoWs are also compared to {\tt Zest}
and also passed in as features, to a Logistic Regression (LR) or Naive Bayes (NB). {\tt Zest} is ablated with 1, 2 and 4 telescoping dictionaries. Additional sentiment analysis datasets (not shown) exhibit similar trends.
\hush{the ensembling {\tt Zest} with BoW slightly outperforms BoW alone.}
}
\end{table}

\subsection{Comparisons with other compression algorithms}
As discussed in Section~\ref{sec:motivation}, Halford \cite{DataCompression} compares various compression algorithms by speed and quality of results on a categorical newsgroups dataset.
While {\tt zstd} is not included, the {\tt LZMA} algorithm performs the best but takes over 30 minutes for 4,000 documents. 
We extend Halford\'s work by including {\tt zstd} in two ways --- as a regular compressor (without using its dictionary mode) and via {\tt Zest}. Such comparisons can clarify if (1) other compression algorithms are competitive and (2) if our improvements help with performance and speed.

Halford uses the Newsgroups dataset \cite{Newsgroups} that contains 20,000 documents from 20 newsgroups. {\tt GZip}, {\tt Zlib}, and {\tt BZ2} all underperform, with an average of 0.65-0.75 for precision and recall (figure-of-merit, or FOM).
The best-performing algorithm\hush{from Halford\'s comparison}, {\tt LZMA}, attains FOM 0.897 and takes 30 minutes to complete. {\tt zstd} used as a regular compressor achieves FOM 0.857 and completes in under a minute. {\tt Zest} takes only 40 seconds to train and test the documents, with FOM 0.899. {\tt Zest} is a significant improvement over using compression algorithms on its own, likely due to the telescoping dictionaries. 

\section{Conclusions and Future Work}

We have demonstrated model-free language-agnostic text features based on data compression that can be useful to text rankers and classifiers. In addition to using a modern data compression tool, our implementation goes beyond the ideas in \cite{LangTrees} by leveraging the {\em dictionary mode} in {\tt zstandard}, using {\em telescoping dictionaries} and performing {\em word padding}. 
Empirical performance of {\tt TorchScript} module on a production ML platform is competitive with that of content-embedding features. However, for some simpler datasets with clear distinction between text classes, BoW shows better ML performance, while being simple and fast.

Evidence from Section~\ref{sec:evaluation}
allows us to conclude the following: 
\begin{enumerate}
\item Compression-based methods can be significantly improved by telescoping dictionaries and word padding.
\item Counting methods can achieve strong performance in some cases.
\item Ability to handle synonyms via embeddings offers no advantage on some practical datasets.
\item Different datasets and classification types favor different models.
\item Zest is competitive with trained sentence embeddings in production settings.
\item Zest outperforms other counting methods such as n-grams in production datasets.
\end{enumerate}

Although {\tt Zest} performance is strong and can be an adequate replacement for content embeddings in a production environment, BoWs as features consistently performed the best on all public datasets that we used to evaluate {\tt Zest}. Cursory analysis suggests that these datasets allow identifying each class by a small set of words,\hush{, which means that counting methods will perform well out of the box.} making explicit counting more accurate than compression-based methods and trained word/content embeddings. However, in a production environment as the one described in Section~\ref{sec:evaluation}, frequent class-specific words are less common, allowing methods like {\tt Zest} to be on par with competition but with no training. We also evaluated (Markov-chain) language models that estimate the probability that a given text was generated by a given language \cite{AItextbook, NLP}. In our experiments (not shown), they produce slightly more accurate classifiers than {\tt Zest}, but tend to be more complex and consume greater resources. Compression-based methods using compression algorithms out of the box do not perform well either, with the exception of {\tt lzma}. {\tt zstd} slightly underperforms in comparison with {\tt lzma}, but can cut the compression time significantly. {\tt Zest} with its telescoping dictionaries is the most efficient and best-performing compression-based method in our experiments.

Our findings are useful when designing ML platforms that need to deal with text features without asking product engineers to write ML code. They suggest maintaining several lightweight, language-agnostic text features including compression-based ones, and letting the model-training process choose which features are helpful. In many applications, such low-hanging features provide performance that is close to or better than more sophisticated word/content embeddings, while using a much smaller resource and latency footprint. Unlike more sophisticated methods, lightweight methods tend to be language agnostic and can be implemented without language detection.

It is straightforward to generalize our methods to the applications explored in \cite{DataMining}, and we believe that many of our conclusions are going to generalize as well.

Our code and experiments are available at \url{https://github.com/facebookresearch/zest}.

\pagebreak

\end{document}